\documentclass{article}
\usepackage{spconf,amsmath,graphicx,amssymb,amsfonts}
\usepackage{booktabs, multicol, multirow, xcolor, threeparttable}
\usepackage{hyperref}
\usepackage{enumitem}
\hypersetup{
    colorlinks=false,
    citebordercolor=[rgb]{0,1,0},
    linkbordercolor=[rgb]{1,0,0},
    linktocpage=true
}
\usepackage{caption}
\usepackage[sort&compress,numbers]{natbib}

\title{WFTNet: Exploiting Global and Local Periodicity in Long-term Time Series Forecasting}
\name{
\hspace{-1.6em}
Peiyuan Liu\textsuperscript{1,2,4}, 
Beiliang Wu\textsuperscript{1,4}, 
Naiqi Li\textsuperscript{2,*}, 
Tao Dai\textsuperscript{1,2,*},
Fengmao Lei\textsuperscript{3}, 
Jigang Bao\textsuperscript{2},
Yong Jiang\textsuperscript{2},
Shu-Tao Xia\textsuperscript{2}
\thanks{Corresponding author: Tao Dai  and Naiqi Li\\
This work is supported in part by National Natural Science Foundation of China, under Grant (62302309, 62171248), Shenzhen Science and Technology Program (Grant No. JCYJ20220818101014030, JCYJ20220818101012025, WDZC20231128114058001), and Swift Fund Fintech Funding 2023.
}}
\address{
$^1$College of Computer Science and Software Engineering, Shenzhen University\\
$^2$Tsinghua Shenzhen International Graduate School, Tsinghua University \\
$^3$Ping An Bank Co., Ltd.
$^4$WeBank Institute of Financial Technology, Shenzhen University
\\
lpy23@mails.tsinghua.edu.cn,  daitao.edu@gmail.com, \{linaiqi, jiangy, xiast\}@sz.tsinghua.edu.cn}
\begin{document}
\maketitle
% \ninept

\begin{abstract}
Recent CNN and Transformer-based models tried to utilize frequency and periodicity information for long-term time series forecasting.
However, most existing work is based on Fourier transform, which cannot capture fine-grained and local frequency structure. In this paper, we propose a Wavelet-Fourier Transform Network (WFTNet) for long-term time series forecasting. WFTNet utilizes both Fourier and wavelet transforms to extract comprehensive temporal-frequency information from the signal, where Fourier transform captures the global periodic patterns and wavelet transform captures the local ones. Furthermore, we introduce a Periodicity-Weighted Coefficient (PWC) to adaptively balance the importance of global and local frequency patterns. Extensive experiments on various time series datasets show that WFTNet consistently outperforms other state-of-the-art baseline. Code is available at \href{https://github.com/Hank0626/WFTNet}{https://github.com/Hank0626/WFTNet}.

\end{abstract}
\section{Introduction}
\label{sec:intro}
Long-term time series forecasting is a crucial task with broad applications in diverse domains, such as financial investment \cite{fjellstrom2022long}, weather prediction \cite{angryk2020multivariate}, and traffic flow estimation \cite{chen2001freeway}. However, due to the inherent complexity of real-world time series data, which often involves intricate temporal variations containing both global and local periodicity, long-term time series forecasting is still a challenging problem.

Recently, Transformer-based models have become increasingly important in time series prediction~\cite{zhou2021informer, wu2021autoformer, liu2021pyraformer, zhou2022fedformer, woo2022etsformer, nie2023time}. However, these methods are insufficient in utilizing temporal information, and have difficulty in capturing intricate periodical patterns~\cite{zeng2023transformers}. To address these challenges, a CNN-based model known as TimesNet has been proposed~\cite{wu2023timesnet}. TimesNet explicitly considers the presence of multiple periodic cycles, and employs Fourier transform to convert the 1D time series into 2D representations to enable the analysis of both intra-period and inter-period variations. 
However, TimesNet primarily emphasizes global periodic structures while often overlooking crucial local periodic patterns.

\begin{figure*}[!t]
  \centering
  \includegraphics[width=\textwidth]{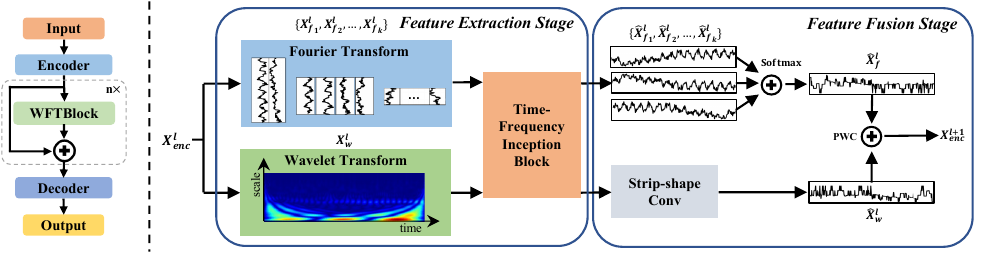}
  \caption{Overall architecture of WFTNet (left) and details of WFTBlock (right). The encoder and decoder manage input normalization, embedding, and output projection. WFTBlocks transform the 1D time series into 2D representations using FFT for global periodic patterns and CWT for local features.}
  \label{fig:framework}
\end{figure*}

Wavelet transform has a unique advantage over Fourier transform in terms of capturing local periodicity in time series data~\cite{zhou2022fedformer, torrence1998practical}. Fourier transform excels at identifying global periodic structures. The key challenge lies in effectively combining their strengths under a unified framework. In this paper, we propose a \textbf{W}avelet-\textbf{F}ourier \textbf{T}ransform \textbf{Net}work (\textbf{WFTNet}), as illustrated in Figure \ref{fig:framework}. WFTNet employs WFTBlock to map the 1D time series into 2D spaces, leveraging both Fourier and wavelet transforms. Specifically, Fourier transform is used to capture global periodicity, and wavelet transform focuses on local periodic patterns. To adaptively balance the importance of global and local periodic patterns, we introduce a Periodicity-Weighted Coefficient (PWC), which measures the relative strength of global periodicity in the signal.

Our contributions are summarized as follows: (1) We propose the WFTNet, a novel model for long-term time series forecasting, which incorporates WFTBlock to effectively capture both global and local periodic patterns in time series data; (2) We introduce the PWC to balance the importance of global and local periodicity output from Fourier and wavelet transforms; and (3) WFTNet achieves consistent state-of-the-art performance on various long-term time series forecasting datasets.

% \ref{fig:framework}. 
\section{Background and Related Work}
\label{sec:format}

\subsection{Discrete Fourier Transform}
\label{ssec:ft}

Discrete Fourier Transform (DFT) \cite{osgood2009fourier} converts a temporal signal from the time domain to the frequency domain. It decomposes a signal into a series of different frequencies. Specifically, given a time series signal $\textbf{x}=[x_{0},...,x_{T-1}]$ of length $T$, $\textbf{x}$ can be transformed as a set of frequency coefficient as:
\begin{equation}
\mathbf{C}_{t}=\sum_{n=0}^{T-1} x_n e^{-j(2 \pi / T) t n} \quad (0 \leq t < T)
\label{eq:fourier}
\end{equation}
where each component $\mathbf{C}_{t}$ is a complex number that contains amplitude and phase information. The DFT is primarily aimed at identifying global periodicity within a signal, rather than capturing local periodic features. For computational efficiency, the Fast Fourier Transform (FFT) is commonly used as a more rapid method to calculate the DFT~\cite{fft}. 

% The amplitude value $\mathbf{A}_{k}$ of each frequency component can be obtained from 
% $\mathbf{A}_{k}=abs(\mathbf{X}_{k})$, which indicates the strength of the corresponding frequency's presence in the original signal. Following the approach outlined before, we can select the frequency with the maximum amplitude, $\mathbf{f}=\frac{1}{p}$, to reshape the original temporal series X$\in$1×T into a two-dimensional form X$\in$n×p. Here, p represents the length of the period, and n corresponds to the number of periods in the original sequence, where $\mathbf{n}=\frac{T}{p}$.

% Viewed from the frequency domain perspective, expanding the input 1D time series into a 2D space can significantly avoid the problems arising from these overlapping and interactive frequencies. Hence, our work will primarily focus on the method to effectively process 1D time series in the frequency domain and further extend them into 2D spaces. In order to enhance the model's ability in representation learning.

\subsection{Continuous Wavelet Transform}
\label{ssec:wt}

Continuous Wavelet Transform (CWT) is another technique to analyze the time-frequency characteristics of a signal~\cite{torrence1998practical, brown2009wavelet}. CWT decomposes signals into both time and frequency domains. Therefore, it is much more effective in capturing local period structures in time series data. Specifically, for a time series signal $\textbf{x} = [x_{0}, ..., x_{T-1}]$, the transformation is defined as:
\begin{equation}
CWT(s, \tau)=\int x_{t} \cdot \frac{1}{\sqrt{s}} \cdot \Psi^*\left(\frac{t-\tau}{s}\right) dt \quad (0 \leq t < T)
\label{eq:wavelet}
\end{equation}
where $s$ represents the scale factor, $\tau$ is the translation factor, $\Psi(t)$ stands for the selected mother wavelet function, and $*$ denotes the complex conjugate operation. The relationship between $\Psi$ and $\Psi^*$ is:
$
\Psi^*_{s, \tau}(t)=\frac{1}{\sqrt{s}} \Psi\left(\frac{t-\tau}{s}\right)
$.

In the field of CWT, a variety of mother wavelet functions are available, such as Haar wavelet, Daubechies wavelet, and Morlet wavelet~\cite{torrence1998practical}. The Morlet wavelet excels in the analysis of time-frequency characteristics in signals~\cite{grossmann1984decomposition}, and is considered as the primary mother wavelet function in our work. It is defined as $\Psi(t)=\pi^{-\frac{1}{4}} e^{j \omega_0 t} e^{-\frac{t^2}{2}}$, where $\omega_{0}$ is the central frequency.
% Assuming that the input is still X, we have chosen a wavelet function $\psi(t)$, that depends on a nondimensional 'time' parameter $t$. For example, the wavelet basis function, Morlet, can be formulated as $\psi(t)=\exp ^{-\frac{t^2}{2}} \cos (5 t)$, then the wavelet transform for input signal X is 

% Finally, we can obtain 2D data in both time and frequency domains.

% That means the Wavelet Transform can perform exceptionally well for non-stationary signals and adapt to different frequency variations at any time. The latter characteristic gives direct information about frequencies that will help the modeling. Consequently, the model will be much more efficient in capturing the instantaneous local representation.
% As a result, it's necessary to consider the advantages of the Fourier Transform and the Wavelet Transform, leveraging their complementary strengths to achieve the model's strong capability in both overall and localized representation learning.

\subsection{Related Work}
\label{ssec:subhead}
CNN and Transformer are commonly used foundational networks in long-term time series forecasting. CNN is good at modeling local features and is applied in models like MICN \cite{wang2023micn}, SCINet \cite{liu2022scinet}, and TimesNet \cite{wu2023timesnet}. Transformer has the ability to capture long-term dependencies, so it is also widely used, such as Informer \cite{zhou2021informer}, FEDformer \cite{zhou2022fedformer}, and PatchTST \cite{nie2023time}.

TimesNet and FEDformer both utilize frequency information. However, FEDformer falls short of fully exploiting periodic patterns in the signal, resulting in less competitive results compared to more recent approaches. Among the existing works, TimesNet represents one of the most well-performing models that applies frequency decomposition. Nonetheless, TimesNet is based on Fourier transform, and thus only captures the global frequency of the entire time series and ignores local frequency variations.

\section{Method}
\label{sec::method}
\subsection{Problem Statement}
The task of long-term time series forecasting starts with a historical sequence $\textbf{X}_{\text{in}}=[\textbf{x}_1,...,\textbf{x}_{T_s}]^\top \in \mathbb{R}^{T_s \times C}$ and aims to predict a future sequence $\textbf{X}_{\text{out}}=[\textbf{x}_{T_s+1},...,\textbf{x}_{T_s+T_p}]^\top \in \mathbb{R}^{T_p \times C}$. Here, $T_s$ and $T_p$ represent the lengths of the past and future time windows, respectively, while $C$ denotes the dimensionality of the time series variables. 

\renewcommand{\arraystretch}{0.95}
\begin{table*}[htb]
\small
\centering
\begin{threeparttable}
\caption{Quantitative evaluation of long-term time series forecasting models with varying output sequence lengths $T\in\{96, 192, 336, 720\}$. Input sequence is fixed at 96. Best performances are highlighted in red and second-best are underlined in blue.}
\begin{tabular}{cccccccccccccc}
\hline
 \multicolumn{2}{c}{Models} & \multicolumn{2}{c}{\textbf{WFTNet}} & \multicolumn{2}{c}{TimesNet \cite{wu2023timesnet}} & \multicolumn{2}{c}{ETSformer \cite{woo2022etsformer}} & \multicolumn{2}{c}{DLinear \cite{zeng2023transformers}} & \multicolumn{2}{c}{FEDformer \cite{zhou2022fedformer}} & \multicolumn{2}{c}{Autoformer \cite{wu2021autoformer}} \\
\cmidrule(lr){3-4} \cmidrule(lr){5-6} \cmidrule(lr){7-8} \cmidrule(lr){9-10}  \cmidrule(lr){11-12} \cmidrule(lr){13-14}   
 \multicolumn{2}{c}{Metric} & MSE & MAE & MSE & MAE & MSE & MAE & MSE & MAE & MSE & MAE & MSE & MAE\\
\hline
\multirow{4}{*}{\rotatebox[origin=c]{90}{ECL}} 

& 96  & \textcolor{red}{\textbf{0.164}} & \textcolor{red}{\textbf{0.267}} & \textcolor[rgb]{0.25, 0.5, 0.75}{\underline{0.167}} & \textcolor[rgb]{0.25, 0.5, 0.75}{\underline{0.271}} & 0.187 & 0.304 & 0.197 & 0.282 & 0.193 & 0.308 & 0.201 & 0.317 \\

& 192 & \textcolor{red}{\textbf{0.181}} & \textcolor{red}{\textbf{0.282}} & \textcolor[rgb]{0.25, 0.5, 0.75}{\underline{0.187}} & \textcolor[rgb]{0.25, 0.5, 0.75}{\underline{0.290}} & 0.199 & 0.315 & 0.196 & 0.285 & 0.201 & 0.315 & 0.222 & 0.334 \\

& 336 & \textcolor{red}{\textbf{0.194}} & \textcolor{red}{\textbf{0.295}} & \textcolor[rgb]{0.25, 0.5, 0.75}{\underline{0.202}} & 0.303 & 0.212 & 0.329 & 0.209 & \textcolor[rgb]{0.25, 0.5, 0.75}{\underline{0.301}} & 0.214 & 0.329 & 0.231 & 0.338 \\

& 720 & \textcolor[rgb]{0.25, 0.5, 0.75}{\underline{0.230}} & \textcolor[rgb]{0.25, 0.5, 0.75}{\underline{0.325}} & \textcolor{red}{\textbf{0.220}} & \textcolor{red}{\textbf{0.318}} & 0.233 & 0.345 & 0.265 & 0.360 & 0.246 & 0.355 & 0.254 & 0.361 \\
\hline
\multirow{4}{*}{\rotatebox[origin=c]{90}{Traffic}} 

& 96  & 0.594 & \textcolor[rgb]{0.25, 0.5, 0.75}{\underline{0.316}} & \textcolor[rgb]{0.25, 0.5, 0.75}{\underline{0.590}} & \textcolor{red}{\textbf{0.314}} & 0.607 & 0.392 & 0.650 & 0.396 & \textcolor{red}{\textbf{0.587}} & 0.366 & 0.613 & 0.388 \\

& 192 & 0.624 & \textcolor[rgb]{0.25, 0.5, 0.75}{\underline{0.332}} & 0.616 & \textcolor{red}{\textbf{0.322}} & 0.621 & 0.399 & \textcolor{red}{\textbf{0.598}} & 0.370 & \textcolor[rgb]{0.25, 0.5, 0.75}{\underline{0.604}} & 0.373 & 0.616 & 0.382 \\

& 336 & 0.631 & \textcolor[rgb]{0.25, 0.5, 0.75}{\underline{0.339}} & 0.634 & \textcolor[rgb]{0.25, 0.5, 0.75}{\underline{0.339}} & 0.622 & 0.396 & \textcolor{red}{\textbf{0.605}} & 0.373 & \textcolor[rgb]{0.25, 0.5, 0.75}{\underline{0.621}} & 0.383 & 0.622 & \textcolor{red}{\textbf{0.337}} \\

& 720 & 0.664 & 0.360 & 0.659 & \textcolor{red}{\textbf{0.349}} & \textcolor[rgb]{0.25, 0.5, 0.75}{\underline{0.632}} & 0.396 & 0.645 & 0.394 & \textcolor{red}{\textbf{0.626}} & \textcolor[rgb]{0.25, 0.5, 0.75}{\underline{0.355}} & 0.660 & 0.408 \\
\hline
\multirow{4}{*}{\rotatebox[origin=c]{90}{Weather}} 

& 96  & \textcolor{red}{\textbf{0.161}} & \textcolor{red}{\textbf{0.210}} & \textcolor[rgb]{0.25, 0.5, 0.75}{\underline{0.169}} & \textcolor[rgb]{0.25, 0.5, 0.75}{\underline{0.219}} & 0.197 & 0.281 & 0.196 & 0.255 & 0.217 & 0.296 & 0.266 & 0.336 \\

& 192 & \textcolor{red}{\textbf{0.211}} & \textcolor{red}{\textbf{0.254}} & \textcolor[rgb]{0.25, 0.5, 0.75}{\underline{0.226}} & \textcolor[rgb]{0.25, 0.5, 0.75}{\underline{0.266}} & 0.237 & 0.312 & 0.237 & 0.312 & 0.276 & 0.336 & 0.307 & 0.367 \\

& 336 & \textcolor{red}{\textbf{0.271}} & \textcolor{red}{\textbf{0.296}} & \textcolor[rgb]{0.25, 0.5, 0.75}{\underline{0.281}} & \textcolor[rgb]{0.25, 0.5, 0.75}{\underline{0.303}} & 0.298 & 0.353 & 0.283 & 0.335 & 0.339 & 0.380 & 0.359 & 0.395 \\

& 720 & \textcolor[rgb]{0.25, 0.5, 0.75}{\underline{0.347}} & \textcolor[rgb]{0.25, 0.5, 0.75}{\underline{0.346}} & 0.357 & 0.353 & 0.352 & \textcolor{red}{\textbf{0.288}} & \textcolor{red}{\textbf{0.345}} & 0.381 & 0.403 & 0.428 & 0.419 & 0.428 \\
\hline
% \multirow{4}{*}{\rotatebox[origin=c]{90}{ETTh1}} 

% & 96  & 0.399 & 0.415 & 0.389 & \textcolor[rgb]{0.25, 0.5, 0.75}{\underline{0.412}} & 0.494 & 0.479 & \textcolor[rgb]{0.25, 0.5, 0.75}{\underline{0.386}} & \textcolor{red}{\textbf{0.400}} & \textcolor{red}{\textbf{0.376}} & 0.419 & 0.449 & 0.459 \\

% & 192 & 0.446 & \textcolor[rgb]{0.25, 0.5, 0.75}{\underline{0.441}} & 0.439 & 0.442 & 0.538 & 0.504 & \textcolor[rgb]{0.25, 0.5, 0.75}{\underline{0.437}} & \textcolor{red}{\textbf{0.432}} & \textcolor{red}{\textbf{0.420}} & 0.448 & 0.500 & 0.482 \\

% & 336 & \textcolor[rgb]{0.25, 0.5, 0.75}{\underline{0.476}} & \textcolor{red}{\textbf{0.454}} & 0.493 & 0.470 & 0.574 & 0.521 & 0.481 & 0.459 & \textcolor{red}{\textbf{0.459}} & \textcolor[rgb]{0.25, 0.5, 0.75}{\underline{0.465}} & 0.521 & 0.496 \\

% & 720 & \textcolor{red}{\textbf{0.482}} & \textcolor{red}{\textbf{0.474}} & 0.517 & \textcolor[rgb]{0.25, 0.5, 0.75}{\underline{0.494}} & 0.562 & 0.535 & 0.519 & 0.516 & \textcolor[rgb]{0.25, 0.5, 0.75}{\underline{0.506}} & 0.507 & 0.514 & 0.512 \\
% \hline
\multirow{4}{*}{\rotatebox[origin=c]{90}{ETT\tnote{*}}} 

& 96  & \textcolor{red}{\textbf{0.323}} & \textcolor{red}{\textbf{0.365}} & \textcolor[rgb]{0.25, 0.5, 0.75}{\underline{0.332}} & \textcolor[rgb]{0.25, 0.5, 0.75}{\underline{0.369}} & 0.340 & 0.391 & 0.333 & 0.387 & 0.358 & 0.397 & 0.346 & 0.388 \\

& 192 & \textcolor[rgb]{0.25, 0.5, 0.75}{\underline{0.403}} & \textcolor{red}{\textbf{0.409}} & \textcolor{red}{\textbf{0.396}} & \textcolor[rgb]{0.25, 0.5, 0.75}{\underline{0.410}} & 0.430 & 0.439 & 0.477 & 0.476 & 0.429 & 0.439 & 0.456 & 0.452 \\

& 336 & \textcolor{red}{\textbf{0.427}} & \textcolor{red}{\textbf{0.433}} & \textcolor[rgb]{0.25, 0.5, 0.75}{\underline{0.446}} & \textcolor[rgb]{0.25, 0.5, 0.75}{\underline{0.447}} & 0.485 & 0.479 & 0.594 & 0.541 & 0.496 & 0.487 & 0.482 & 0.486 \\

& 720 & \textcolor{red}{\textbf{0.430}} & \textcolor{red}{\textbf{0.445}} & \textcolor[rgb]{0.25, 0.5, 0.75}{\underline{0.434}} & \textcolor[rgb]{0.25, 0.5, 0.75}{\underline{0.448}} & 0.500 & 0.497 & 0.831 & 0.657 & 0.463 & 0.474 & 0.515 & 0.511 \\
\hline

\multicolumn{2}{c}{$1^{st}$ Count} & \multicolumn{2}{c}{\textcolor{red}{\textbf{19}}} & \multicolumn{2}{c}{\textcolor[rgb]{0.25, 0.5, 0.75}{\underline{6}}} & \multicolumn{2}{c}{1} & \multicolumn{2}{c}{3} & \multicolumn{2}{c}{2} & \multicolumn{2}{c}{1} \\
\hline
\end{tabular}
\begin{tablenotes}
    \footnotesize
    \item[*] ETT means the ETTh2. Experiments were also conducted on ETTm1, ETTm2, and ETTh1 datasets, but are omitted here due to space constraints.
\end{tablenotes}
\label{tab::res}
\end{threeparttable}
\end{table*}

\subsection{Framework}
WFTNet employs an encoder-decoder framework augmented by several Wavelet-Fourier Transform Blocks (WFTBlocks) in a residual way \cite{he2016deep}. The encoder first normalizes the input matrix \( \textbf{X}_{\text{in}} \) to produce \( \textbf{X}_{\text{norm}} \). This normalized data is then transformed by the Data Embedding process into the feature space \( \textbf{X}_{\text{enc}} \in \mathbb{R}^{T_e \times D} \), where \( D \) representing the embedding dimension and \( T_e = T_s + T_p \). This embedding technique synthesizes value and positional encodings while applying dropout regularization to mitigate overfitting. After the process of encoding, the data moves through several WFTBlocks, which transform the 1D time series into 2D spatial representations using FFT and CWT. These blocks utilize efficient convolutional networks to effectively capture both local and global periodic patterns. During the decoding stage, \( \textbf{X}_{\text{enc}} \) undergoes a linear projection to produce the output time series window \( \textbf{X}_{\text{out}} \). A final de-normalization step yields the forecasted time series. In general, WFTNet offers a comprehensive approach to predicting multivariate time series data by utilizing the combined benefits of wavelet and Fourier transforms, convolutional structures, and advanced encoding-decoding techniques.

\subsection{WFTBlock}
The WFTBlock aims to perform time-frequency feature extraction from an input 1D time series \( \textbf{X}^{l}_{\text{enc}}\), where $\textbf{X}^{l}_{\text{enc}}$ represents the input to the $(l+1)^{th}$ WFTBlock. The architecture of WFTBlock is divided into two main stages: the Feature Extraction Stage and the Feature Fusion Stage.

\textbf{Feature Extraction Stage:} The input time series \( \textbf{X}^{l}_{\text{enc}}\) is splitted into two distinct branches: one subjected to Fourier transform and the other to wavelet transform. 

The Fourier transform branch applies FFT (Eq. \ref{eq:fourier})  to produce the amplitude \( \textbf{a}=[a_1,\ldots, a_{T_e}] \), where $a_i=|C_i|$. The top \( k \) frequencies \( [f_1,\ldots, f_k] \) with the highest amplitudes \( [a_1,\ldots, a_k] \) are selected to yield corresponding periods \( [p_1,\ldots, p_k] \), where \( p_i = \left\lceil \frac{1}{f_i} \right\rceil \). Given \( T_e \) and a selected period \( p_i \), a 2D frequency map \( x_{f_i}^{l} \) is generated by segmenting the original sequence into blocks of length \( p_i \) and stacking them column-wise until \( T_e \) elements are covered. Zero-padding is applied to complete the last column if \( T_e / p_i \) is not an integer~\cite{wu2023timesnet}. 

The wavelet transform branch employs CWT (Eq. \ref{eq:wavelet}) to produce a time-frequency map \(\textbf{X}^l_w \in \mathbb{R}^{T_e\times s}\), where \( s \) denotes the wavelet scale. This map provides superior time-frequency localization compared to the 2D frequency map derived from Fourier transform. While FFT provides a global view of frequencies, the multi-scale nature of \(\textbf{X}^l_w\) enables precise localization of frequency components in both time and frequency domains. This attribute is particularly beneficial for analyzing non-stationary time series, where frequency components and their corresponding periods can vary dynamically over time.

The outputs from both branches are further processed by a Time-Frequency Inception Block \cite{szegedy2017inception} to harvest global and local periodic features.

\textbf{Feature Fusion Stage:} At this stage, the transformed outputs from both the Fourier and wavelet branches are combined. In the Fourier branch, each frequency component \( \hat{\textbf{X}}_{f_i}^{l} \) is weighted according to its corresponding amplitude \( a_i \) using a softmax normalization:
$
\hat{\textbf{X}}_f^{l} = \sum_{i=1}^{k} \text{Softmax}(a_i) \times \hat{\textbf{X}}_{f_i}^{l}
$.
This weighted sum essentially provides an importance-adjusted composite frequency representation of the original sequence. For the wavelet branch, a specialized strip-shaped convolutional kernel is applied to the output to compress the scale dimension $s$, resulting in \( \hat{\textbf{X}}_w^{l} \). 

{
\captionsetup[figure]{width=0.9\linewidth}
\begin{figure}[htb]
    \centering
    \includegraphics[width=\linewidth]{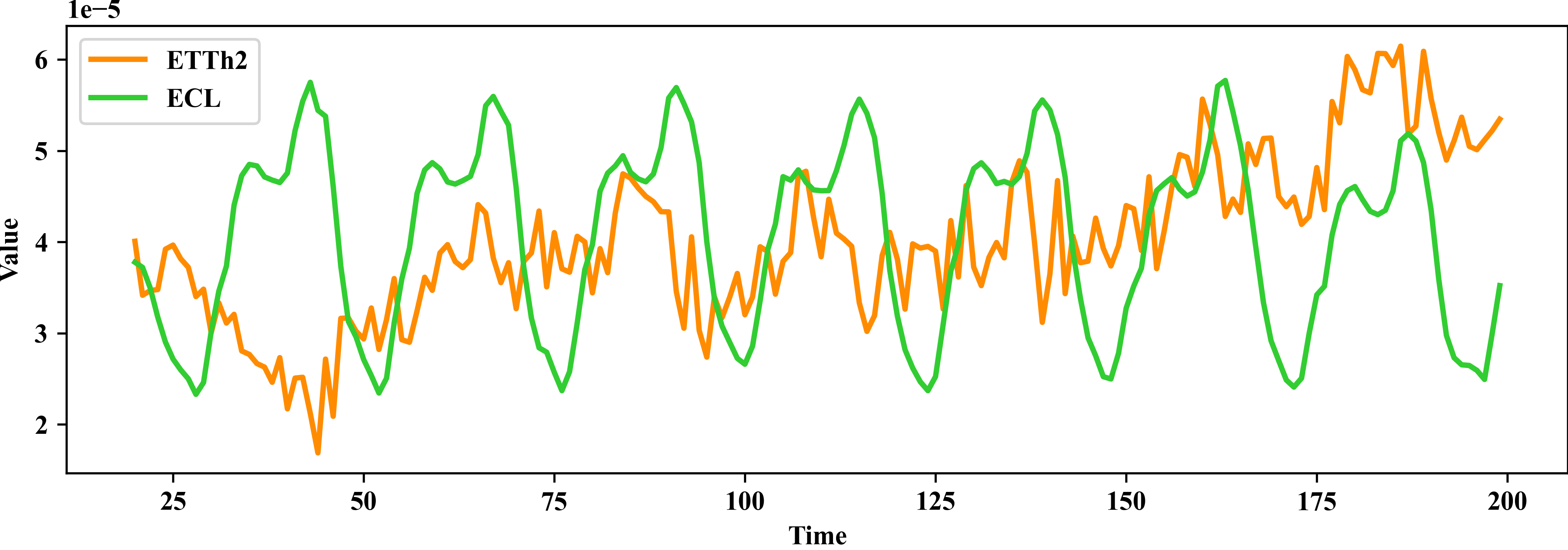}
    \caption{Visualization of the normalized mean channel values for the ECL and ETTh2 datasets. The plot reveals stronger periodicity in ECL compared to ETTh2.}
    \label{fig:period}
\end{figure}
}

The weighted outputs from both branches are then combined using the Periodicity-Weighted Coefficient (PWC) \( \alpha \) by:
\[
\textbf{X}^{l+1}_{\text{enc}} = \alpha^n \hat{\textbf{X}}_f^{l} + (1 - \alpha^n) \hat{\textbf{X}}_w^{l}
\]
where \( n \) is a hyperparameter that modulates the contribution of $\alpha$.

\subsection{Periodicity-Weighted Coefficient}
The Periodicity-Weighted Coefficient (PWC), denoted by \( \alpha \), adaptively balances Fourier and wavelet transformations for global and local periodicities, respectively. \( \alpha \) quantifies inherent periodicity to optimally weight the contributions from each transform method. To calculate \( \alpha \), we perform a Fourier transform on each of the \( C \) channels in the time series, and determine the average ratio of the maximum to total energy across these channels within the first \( m \) frequencies:
\[
\alpha = \frac{1}{C} \sum_{i=1}^{C} \frac{\max \left( a_i^2 \right)_{i=1,\ldots,m}}{\sum_{i=1}^{m} a_i^2}
\]
where $a_i$ represents the amplitude of the $i^{th}$ frequency after Fourier transform.

The adaptability of \( \alpha \) makes it effective for diverse time series. A value close to 1 emphasizes global periodicity via Fourier features, while a value near 0 focuses on localized behavior through wavelet features.

\section{Experiments}
In this section, we assess the performance of WFTNet using Mean Squared Error (MSE) and Mean Absolute Error (MAE) as key performance metrics, in line with prior research~\cite{wu2021autoformer, zhou2022fedformer, zeng2023transformers, woo2022etsformer, wu2023timesnet}.

\subsection{Datasets}
To rigorously validate our method, we conduct experiments on seven benchmark time series datasets: Electricity Transformer Temperature (ETT) with its four sub-datasets (ETTh1, ETTh2, ETTm1, ETTm2)~\cite{zhou2021informer}, Traffic, ECL, and Weather datasets~\cite{lai2018modeling}. For each dataset, we allocate 70\% for training, 20\% for testing, and the remaining 10\% for validation.

\begin{table}[htb]
\small
\centering
\begin{threeparttable}
\caption{Comparison of the performance of WFTNet with its Fourier-Only and Wavelet-Only variants.}
\begin{tabular}{cccccccccccccc}
\hline
 \multicolumn{2}{c}{Models} & \multicolumn{2}{c}{WFTNet} & \multicolumn{2}{c}{Fourier-Only} & \multicolumn{2}{c}{Wavelet-Only} \\
\cmidrule(lr){3-4} \cmidrule(lr){5-6} \cmidrule(lr){7-8} 
 \multicolumn{2}{c}{Metric} & MSE & MAE & MSE & MAE & MSE & MAE \\
\hline
 \multirow{4}{*}{\rotatebox[origin=c]{90}{ECL}} 
 & 96  & \textcolor{red}{\textbf{0.164}} & \textcolor{red}{\textbf{0.267}} & \textcolor[rgb]{0.25, 0.5, 0.75}{\underline{0.168}} & \textcolor[rgb]{0.25, 0.5, 0.75}{\underline{0.273}} & 0.196 & 0.301 \\
 & 192 & \textcolor{red}{\textbf{0.181}} & \textcolor{red}{\textbf{0.282}} & \textcolor[rgb]{0.25, 0.5, 0.75}{\underline{0.187}} & \textcolor[rgb]{0.25, 0.5, 0.75}{\underline{0.290}} & 0.209 & 0.309 \\
 & 336 & \textcolor{red}{\textbf{0.194}} & \textcolor{red}{\textbf{0.295}} & \textcolor[rgb]{0.25, 0.5, 0.75}{\underline{0.201}} & \textcolor[rgb]{0.25, 0.5, 0.75}{\underline{0.300}} & 0.217 & 0.318 \\
 & 720 & \textcolor[rgb]{0.25, 0.5, 0.75}{\underline{0.230}} & \textcolor[rgb]{0.25, 0.5, 0.75}{\underline{0.325}} & \textcolor{red}{\textbf{0.218}} & \textcolor{red}{\textbf{0.320}} & 0.247 & 0.347 \\
 \hline
 \multirow{4}{*}{\rotatebox[origin=c]{90}{ETTh2}} 
 & 96  & \textcolor{red}{\textbf{0.323}} & \textcolor[rgb]{0.25, 0.5, 0.75}{\underline{0.365}} & 0.332 & 0.369 & \textcolor[rgb]{0.25, 0.5, 0.75}{\underline{0.329}} & \textcolor{red}{\textbf{0.362}} \\
 & 192 & \textcolor{red}{\textbf{0.403}} & \textcolor{red}{\textbf{0.409}} & 0.406 & 0.412 & \textcolor[rgb]{0.25, 0.5, 0.75}{\underline{0.404}} & \textcolor[rgb]{0.25, 0.5, 0.75}{\underline{0.410}} \\
 & 336 & \textcolor{red}{\textbf{0.427}} & \textcolor{red}{\textbf{0.433}} & 0.446 & 0.447 & \textcolor[rgb]{0.25, 0.5, 0.75}{\underline{0.433}} & \textcolor[rgb]{0.25, 0.5, 0.75}{\underline{0.437}} \\
 & 720 & \textcolor[rgb]{0.25, 0.5, 0.75}{\underline{0.430}} & \textcolor[rgb]{0.25, 0.5, 0.75}{\underline{0.445}} & 0.434 & 0.448 & \textcolor{red}{\textbf{0.421}} & \textcolor{red}{\textbf{0.439}} \\
\hline
\end{tabular}
\label{tab::aba_res}
\end{threeparttable}
\end{table}

\subsection{Main Results}
As evidenced in Table \ref{tab::res}, WFTNet is thoroughly evaluated against other baseline methods, including TimesNet \cite{wu2023timesnet}, ETSformer \cite{woo2022etsformer}, DLinear \cite{zeng2023transformers}, FEDformer \cite{zhou2022fedformer}, and Autoformer \cite{wu2021autoformer}. Our model consistently outperforms these established approaches over varying output sequence lengths, underlining its effectiveness in long-term time series forecasting. These empirical findings substantiate WFTNet's unique strengths compared to existing techniques.

\subsection{Significance of PWC}
The ablation study presented in Table \ref{tab::aba_res} emphasizes the critical role of PWC in WFTNet. In this example, we consider two datasets: ECL and ETTh2, which are shown in Figure \ref{fig:period}. We can clearly see that ECL has stronger periodicity than ETTh2. For clarity, ``Fourier-Only'' refers to the variant of WFTNet where only the Fourier transform branch is activated, while "Wavelet-Only" uses just the wavelet branch. These isolated branches serve as specialized baselines for comparison.  The Fourier branch is particularly advantageous for the ECL, while the Wavelet branch proves more beneficial for the less periodic ETTh2 dataset. Notably, by leveraging PWC for dynamic feature balancing, WFTNet consistently outperforms these specialized branches in both MSE and MAE across two datasets, thereby validating the capability of PWC to seamlessly enhance model accuracy.

\section{Conclusion}
In this paper, we introduce WFTNet for long-term time series forecasting. Leveraging the strengths of Fourier and wavelet transforms via WFTBlock, WFTNet can simultaneously capture global and local period structures of time series data. The proposed Periodicity-Weighted Coefficient adaptively balances these features,  further improving the model’s performance on datasets with various characteristics. Finally, the superiority of WFTNet is confirmed through extensive evaluations, demonstrating its effectiveness and robustness.

\vfill\pagebreak

% \section{REFERENCES}
% \label{sec:refs}

% List and number all bibliographical references at the end of the
% paper. The references can be numbered in alphabetic order or in
% order of appearance in the document. When referring to them in
% the text, type the corresponding reference number in square
% brackets as shown at the end of this sentence \cite{C2}. An
% additional final page (the fifth page, in most cases) is
% allowed, but must contain only references to the prior
% literature.

\bibliographystyle{IEEEbib}
\bibliography{timesnet, patchtst, etsformer, dlinear, fedformer, autoformer, resnet, informer, micn, modeling, wavelet_guide, Fourier_application, wavelet, wavelet_signal, fft, Inception_Block_V4, scinet, pyraformer, weather, traffic_flow, financial}

\end{document}